\documentclass{article}
\usepackage{arxiv-one_col}
\PassOptionsToPackage{table,dvipsnames}{xcolor}  

\usepackage[utf8]{inputenc} 
\usepackage[T1]{fontenc}    
\usepackage{hyperref}       
\usepackage{url}            
\usepackage{amsfonts}       
\usepackage{nicefrac}       
\usepackage{graphicx}
\usepackage[numbers]{natbib}
\usepackage{multirow}
\usepackage{tabularx}
\usepackage{ragged2e}
\usepackage{enumitem}
\usepackage{booktabs} 
\usepackage{threeparttable} 
\usepackage{xtab}
\usepackage{longtable}
\usepackage{orcidlink}
\usepackage{quoting}
\usepackage{subcaption}
\usepackage{array}
\usepackage{ragged2e}
\usepackage{amsmath} 
\usepackage{colortbl} 

\newcommand{\lightrule}{\arrayrulecolor{gray!60}\midrule[0.1pt]\arrayrulecolor{black}}

\hypersetup{
    colorlinks=true,
    linkcolor=blue,
    citecolor=teal,      
    urlcolor=blue,
}

\definecolor{myblue}{RGB}{0, 102, 204}
\definecolor{myorange}{RGB}{204, 102, 0}
\definecolor{mygreen}{RGB}{0, 153, 0}
\definecolor{mypurple}{RGB}{102, 0, 102}

\title{EduRABSA: An Education Review Dataset for Aspect-based Sentiment Analysis Tasks
}

\author{
  Yan Cathy Hua \orcidlink{0000-0001-9155-9667} \qquad
  Paul Denny \orcidlink{0000-0002-5150-9806} \qquad
  J{\"o}rg Wicker \orcidlink{0000-0003-0533-3368}  \qquad
  Katerina Taskova \orcidlink{0000-0002-3217-7877} \\
  \\
  School of Computer Science, University of Auckland, New Zealand
}

\date{}

\renewcommand{\headeright}{}
\renewcommand{\undertitle}{}
\renewcommand{\shorttitle}{EduRABSA: An Education Review Dataset for Aspect-based Sentiment Analysis Tasks}

\hypersetup{
pdftitle={EduRABSA - An Education Review Dataset for Aspect-based Sentiment Analysis Tasks},
pdfauthor={Yan (Cathy) ~Hua, Paul ~Denny, J{\"o}rg ~Wicker, Katerina ~Taskova},
pdfkeywords={dataset, annotation tool, ABSA, aspect-based sentiment analysis, education domain, student feedback},
}

\begin{document}
\setlength{\leftmargini}{1em} 
\maketitle

\begin{abstract} 

Every year, most educational institutions seek and receive an enormous volume of text feedback from students on courses, teaching, and overall experience. Yet, turning this raw feedback into useful insights is far from straightforward.  It has been a long-standing challenge to adopt automatic opinion mining solutions for such education review text data due to the content complexity and low-granularity reporting requirements. Aspect-based Sentiment Analysis (ABSA) offers a promising solution with its rich, sub-sentence-level opinion mining capabilities. However, existing ABSA research and resources are very heavily focused on the commercial domain. In education, they are scarce and hard to develop due to limited public datasets and strict data protection. A high-quality, annotated dataset is urgently needed to advance research in this under-resourced area. In this work, we present \textsc{EduRABSA} (Education Review ABSA), the first public, annotated ABSA education review dataset that covers three review subject types (course, teaching staff, university) in the English language and all main ABSA tasks, including the under-explored implicit aspect and implicit opinion extraction. We also share \textsc{ASQE-DPT} (Data Processing Tool), an offline, lightweight, installation-free manual data annotation tool that generates labelled datasets for comprehensive ABSA tasks from a single-task annotation. Together, these resources contribute to the ABSA community and education domain by removing the dataset barrier, supporting research transparency and reproducibility, and enabling the creation and sharing of further resources. The dataset, annotation tool, and scripts and statistics for dataset processing and sampling are available at \url{https://github.com/yhua219/edurabsa_dataset_and_annotation_tool}.

\end{abstract}



\section{Introduction}\label{sec1_intro}


Aspect-based Sentiment Analysis (ABSA) is a type of fine-grained sentiment analysis that identifies \textbf{opinions} and their target entities or their attributes (``\textbf{aspects}'') at the sub-sentence level, and classifies them into \textbf{categories} and/or \textbf{sentiment polarities} \cite{hua2024absa, batch2_survey_absa}. An ABSA problem comprises one or more subtasks that output various combinations of these components as shown in Table \ref{table_absa_example}. This low granularity and the ability to turn varied raw expressions into standardised category and sentiment labels make ABSA an ideal choice for mining targeted insights from nuanced opinionated text \cite{batch2_survey_absa, batch2_survey_absadl}. It has been applied to mine online reviews and social media posts across domains, from product/services to public policies, news, and healthcare \cite{hua2024absa}. However, distinguishing different and relevant opinions and aspects within a sentence relies heavily on the textual context and domain knowledge. This makes ABSA particularly domain-dependent and thus resource-sensitive\cite{hua2024absa, batch2_survey_absa}.

\renewcommand{\arraystretch}{1.3}

\begin{table*}[!htbp]
\centering
\footnotesize
\caption{Example of ABSA components and subtask outputs}
\label{table_absa_example}

\begin{tabularx}{\linewidth}{
>{\raggedright\arraybackslash}p{0.25\linewidth} 
>{\raggedright\arraybackslash}p{0.15\linewidth} 
>{\raggedright\arraybackslash}p{0.15\linewidth} 
>{\raggedright\arraybackslash}p{0.15\linewidth}  
>{\raggedright\arraybackslash}p{0.15\linewidth}
}
\toprule

\textbf{Example text} & \multicolumn{4}{l}{\textit{It’s expensive but the pizza is good.}} \\ \midrule
\textbf{ABSA Component} & \textbf{Aspect} & \textbf{Opinion} & \textbf{Category} & \textbf{Sentiment} \\

\textbf{Relation group 1} & (implicit) \textsuperscript{\textcolor{myblue}{1}} & \textcolor{myorange}{expensive} & \textcolor{mygreen}{price} & \textcolor{mypurple}{neutral} \\
\textbf{Relation group 2} & \textcolor{myblue}{pizza} & \textcolor{myorange}{good} & \textcolor{mygreen}{food} & \textcolor{mypurple}{positive} \\
\midrule

\multicolumn{2}{l}{\textbf{ABSA Tasks}} & \multicolumn{2}{l}{\textbf{Output}} \\

\multicolumn{2}{l}{Aspect Extraction (AE)} & \multicolumn{3}{l}{{[}\textcolor{myblue}{null}, \textcolor{myblue}{pizza}{]}} \\
\multicolumn{2}{l}{Opinion Extraction (OE)} & \multicolumn{3}{l}{{[}\textcolor{myorange}{expensive}, \textcolor{myorange}{good}{]}} \\
\multicolumn{2}{l}{Aspect Category Detection (ACD)} & \multicolumn{3}{l}{{[}\textless{}\textcolor{myblue}{null}, \textcolor{mygreen}{price}\textgreater{}, \textless{}\textcolor{myblue}{pizza}, \textcolor{mygreen}{food}\textgreater{}{]}} \\
\multicolumn{2}{l}{Aspect Sentiment Classification (ASC)} & \multicolumn{3}{l}{{[}\textless{}\textcolor{myblue}{null}, \textcolor{mypurple}{neutral}\textgreater{}, \textless{}\textcolor{myblue}{pizza}, \textcolor{mypurple}{positive}\textgreater{}{]}} \\
\multicolumn{2}{l}{Aspect-Opinion Pair Extraction (AOPE)} & \multicolumn{3}{l}{{[}\textless{}\textcolor{myblue}{null}, \textcolor{myorange}{expensive}\textgreater{}, \textless{}\textcolor{myblue}{pizza}, \textcolor{myorange}{good}\textgreater{}{]}} \\
\multicolumn{2}{l}{Aspect Sentiment Triplet Extraction (ASTE) \textsuperscript{\textcolor{myblue}{2}}} & \multicolumn{3}{l}{{[}\textless{}\textcolor{myblue}{null}, \textcolor{myorange}{expensive}, \textcolor{mypurple}{neutral}\textgreater{}, \textless{}\textcolor{myblue}{pizza}, \textcolor{myorange}{good}, \textcolor{mypurple}{positive}\textgreater{}{]}} \\
\multicolumn{2}{l}{Aspect-Sentiment Quadruplet Extraction (ASQE) \textsuperscript{\textcolor{myblue}{3}}} & \multicolumn{3}{l}{{[}\textless{}\textcolor{myblue}{null}, \textcolor{myorange}{expensive}, \textcolor{mygreen}{price}, \textcolor{mypurple}{neutral}\textgreater{}, \textless{}\textcolor{myblue}{pizza}, \textcolor{myorange}{good}, \textcolor{mygreen}{food}, \textcolor{mypurple}{positive}\textgreater{}{]}} \\

\bottomrule
\end{tabularx}

\vspace{1ex}
\caption*{
\begin{minipage}{\textwidth}
\small
\setlength{\baselineskip}{1.2\baselineskip}
\hangindent=2.5em \hangafter=1
\textsuperscript{1} This is an implicit aspect that is absent from the text (marked `null') but its category and relations can be inferred from the context. \par
\hangindent=2.5em \hangafter=1
\textsuperscript{2} Despite the task name, each ASTE triplet contains an aspect term, an opinion term, and a sentiment label. \par
\hangindent=0.75em \hangafter=1
\textsuperscript{3} Despite the task name, each ASQE quadruplet contains an aspect term, an opinion term, a category label, and a sentiment label. \par
\end{minipage}
}

\end{table*}
\renewcommand{\arraystretch}{1}

Despite the rapid growth of ABSA research in the last decade, the availability of public dataset resources has remained heavily concentrated in the commercial product and service review domain, while important domains such as healthcare and education are disproportionately under-represented \cite{hua2024absa}. In a recent systematic review of ABSA studies published between 2008 and 2023 \cite{hua2024absa}, 
education review emerged as the third most common domain, yet it was represented by only 12 primary studies (2.31\%) out of the 519 reviewed.
 
The situation has not substantially improved -- we replicated the search for peer-reviewed literature from 2024 to mid-2025 and identified only three additional studies in this domain.
Together, these 15 education review ABSA studies highlight a shortage of public datasets in this domain: as shown in Table \ref{table_eduabsa_studies}, 12 of them created datasets from scratch, nine went through manual annotation, and only two offer the labelled datasets upon request. 

The lack of public ABSA datasets reflects a primary challenge with education reviews: such data often comes from course and teaching evaluations or student surveys protected by institutional data rules, and thus cannot be publicly shared. Some institutions also restrict third-party tools or servers allowed to access this data, limiting the data-handling choice and capability. 
This shortage of public datasets has slowed the education sector's adoption of ABSA -- a solution well suited to a persistent challenge: making sense of the vast volume of student review text. Institutions collect this feedback regularly, but struggle to extract timely, actionable insights to inform reporting and intervention.

Student reviews vary widely in formality, length, and complexity. A single review or sentence can touch on multiple subtopics, making review-level analysis difficult. 
However, institutions often wish to analyse student feedback on specific aspects (e.g. course organisation, assessments, teaching quality) and aggregate or slice them across levels and dimensions (e.g. by courses, departments, programmes) and join with student or course attributes. This requires a solution that can identify target aspects and related opinions within review entries, and also standardise them for further aggregation and analysis  \cite{edu_8_uae, edu_10_ontology}. These objectives are exactly what ABSA offers through its subtasks \cite{hua2024absa, batch2_survey_absa}: parsing through text to extract relevant opinion and aspect expressions within each sentence (AE, OE, AOPE), and assigning them pre-defined category- and/or sentiment-labels (ACD, ASC, ASTE, ASQE). The ABSA output is at the sub-sentence level and linked to each review entry, and the category and sentiment labels can be easily aggregated and analysed. 

In reality, few educational institutions have the resources to develop their own ABSA datasets, models, or evaluation benchmarks for assessing the quality of external ABSA tools when allowed by the data policy. As a result, student review texts are often under-analysed and under-reported due to the huge volume and the lack of an appropriate automated solution \cite{edu_1_hmm, edu_5_autoscoring}.  Given the domain-dependence and the dominance of supervised training methods among existing ABSA studies \cite{hua2024absa}, having a public ABSA-annotated student review dataset is thus essential for enabling the creation of more public resources and models in this domain. 

\subsection{Our Work and Contributions}

We address the ABSA dataset shortage in the education review domain by presenting \textbf{\textsc{EduRABSA} (Education Review ABSA)}, a collection of 6,500 entries of real student reviews in the English language on courses, teaching staff, and university. The dataset was manually annotated and covers all main ABSA tasks, including challenging and less studied ones. 

We also share \textbf{\textsc{ASQE-DPT}}, an offline, installation-free, lightweight ABSA data annotation tool. It supports manual annotation of all main ABSA tasks, and can automatically parse ASQE and ASTE annotations into other task dataset files. This tool is simple to use and is suitable for all ABSA domains, particularly for processing protected data or for low-resource environments.

\begin{figure*}[!htbp]
  \centering
  \includegraphics[width=0.9\linewidth]{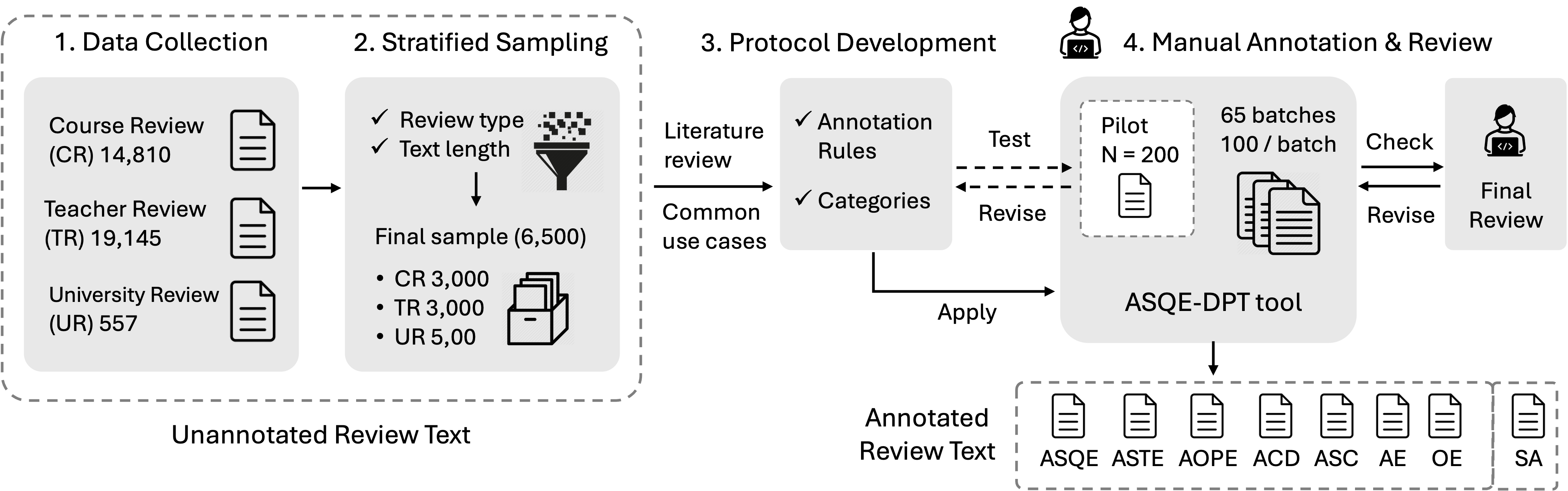}
  \caption{The overall workflow of creating the EduRABSA dataset}
  \label{fig_overall_image}
\end{figure*}

\newpage
Our resources make the following contributions and impact:

\begin{enumerate}

    \item \textbf{Uniqueness and Necessity: } To our knowledge, EduRABSA is the first public dataset of student reviews on all three subjects (course, teaching staff, and university), including annotations for comprehensive ABSA tasks. As Table \ref{table_dataset_stats} shows, it has 7.8 times more reviews and 4.5 times more sentences than the three SemEval 2015 ABSA benchmarks datasets combined \cite{semeval2015}. It fills a long-standing resource gap in the ABSA community, enables research advancement in this under-studied domain, and contributes to the collaboration, transparency, and reproducibility of education review ABSA research by removing the data barrier. 

    \item \textbf{Dataset Usefulness:} \quad  The EduRABSA dataset annotates all ABSA components and their relationships, and thus can support model training and evaluation of all main ABSA tasks. This includes AE and OE for both explicit and implicit aspects and opinions, ACD, ASC, and the composite tasks AOPE, ASTE, ASQE, and the overall review-level sentiment classification (SA). Its task range supports model training and evaluation with single- and multi-task approaches, under the pipeline and end-to-end frameworks \cite{batch2_survey_absa, jointabsa, hua2024absa, datasetsurvey2023}. As the majority of existing ABSA approaches are supervised \cite{hua2024absa}, it can also be used for data augmentation and enables more diverse ABSA approaches such as semi-supervised learning and contrastive learning (e.g. \cite{ref_2025_04_contrastive, implicitOE_2}). The dataset is machine-learning ready and is presented in multiple formats, including that for the popular PyABSA \cite{pyabsa} platform. 

    \item \textbf{Source Data Authenticity:} \quad The dataset was sourced from public-domain licensed, unannotated student review text. The reviews have a low risk of being machine-generated as the majority (6,000 out of 6,500) of the original entries were published before the prevalence of Generative Large Language Models (LLMs).

    \item \textbf{Robustness:} \quad The dataset supports the development of robust models by incorporating realistic complexity and challenging ABSA tasks. It was designed based on our observations of real-world student reviews to reflect a comparable range of length, language style, and content complexity, featuring multiple aspects and sentiments in most sentences. These features are illustrated in Tables~\ref{table_benchmark_dataset_comparison} and \ref{table_dataset_text_length}: compared with the main benchmark ASQE datasets \cite{compound_task_survey_2025}, our dataset shows much higher per-review ratios of quadruplets and unique sub-components, and notably longer aspect and opinion expressions, with the majority spanning up to four times those in existing benchmarks. The annotation captures the prevalent yet largely overlooked implicit opinion expression, which will enable better model generalisation  \cite{implicitOE_2, implicitOE_4, hua2024absa}.

    \item \textbf{Reproducibility :} \quad  To ensure reproducibility and encourage collaboration and further development, we also share the sampling script, sample distribution analysis code, annotation protocol and notes, dataset format conversion script, and raw annotation file.  

    \item \textbf{Enabling Further Resource Creation:} The ASQE-DPT tool is highly mobile and lowers the barrier of data annotation with its installation-free, single-file, browser-based design. It is a useful tool for low-resource settings and a safe offline choice for protected data. It also supports annotation checking and modification, and thus contributes to the improvement and adaptation of existing ABSA datasets. For example, the raw annotation files we share can be directly uploaded to this tool for further quality improvement or category changes. 
    
\end{enumerate}

Our dataset, annotation tool, and the source data and scripts mentioned above are available at \url{https://github.com/yhua219/edurabsa_dataset_and_annotation_tool}.

\renewcommand{\arraystretch}{1.2}

\begin{table}[!htbp]
\centering
\footnotesize

\caption{EduRABSA dataset overview - number of review text entries and extracted quadruplets by review type and sentiment category. For reference, the number of reviews and sentences in the SemEval 2015 ABSA dataset is: Laptop (450, 2500), Restaurant (350, 2000), Hotels (30, 266) \cite{semeval2014}.} \vspace{3pt}

\label{table_dataset_stats}

\begin{tabularx}{1\textwidth}{
>{\raggedright\arraybackslash}p{0.15\linewidth} |
>{\raggedleft\arraybackslash}p{0.08\linewidth} 
>{\raggedleft\arraybackslash}p{0.1\linewidth} 
>{\raggedleft\arraybackslash}p{0.12\linewidth} |
>{\raggedleft\arraybackslash}p{0.1\linewidth}  
>{\raggedleft\arraybackslash}p{0.1\linewidth}
>{\raggedleft\arraybackslash}p{0.1\linewidth}
}

\toprule
\textbf{Review Type} & \textbf{Entries} & \textbf{Sentences} & \textbf{Quadruplets} & \textbf{Positive} & \textbf{Neutral} & \textbf{Negative}\textsuperscript{\textcolor{blue}{1}}
\\ 

\midrule

course review & 3,000 & 8,689 & 10,816 & 5,093 & 1,624 & 4,099 \\
teacher review & 3,000 & 11,576 & 14,530 & 9,102 & 1,066 & 4,362 \\
university review & 500 & 1,011 & 1,691 & 1,380 & 72 & 239 \\
\lightrule
\textbf{TOTAL} & \textbf{6,500} & \textbf{21,276} & \textbf{27,037} & \textbf{15,575} & \textbf{2,762} & \textbf{8,700} \\ 

\bottomrule
\end{tabularx}

\caption*{
\begin{minipage}{\textwidth}
\footnotesize
\setlength{\baselineskip}{1.2\baselineskip}
\hangindent=1em \hangafter=1 \hspace{0.05\textwidth} 
\textsuperscript{1} The last three columns show the number of quadruplets per each sentiment category. \par
\end{minipage}
}

\end{table}

\renewcommand{\arraystretch}{1}


\renewcommand{\arraystretch}{1.2}

\begin{table}[!htbp]
\centering
\footnotesize

\caption{Detailed comparison - EduRABSA vs. benchmark ASQE datasets. For each dataset, the top row reports 1) the number of review entries by dataset split (columns 2--4), and 2) the total number of unique components per quadruplet (columns 5--8). The second row presents quadruplet counts by dataset split and sentiment polarity, where (+), (o), and (–) denote quadruplets with positive, neutral, and negative sentiment labels, respectively. Detailed statistics for the benchmark datasets are taken from \cite{compound_task_survey_2025}. } \vspace{3pt}

\label{table_benchmark_dataset_comparison}

\resizebox{1\linewidth}{!}{%
\begin{tabular}{l | ccc | rrrr}

\toprule
\textbf{Dataset} & \textbf{Train} & \textbf{Val} & \textbf{Test} & \textbf{Aspect} & \textbf{Opinion} & \textbf{Category} & \textbf{Sentiment} \\
 & (+, o, -) & (+, o, -) & (+, o, -) &  &  &  &  \\ 
\midrule

\textbf{ASQP Rest15} & 834 & 209 & 537 & 2125 & 2304 & 2071 & 1699 \\
 & (1005, 34, 315) & (252, 14, 81) & (453, 37, 305) &  &  &  &  \\ 
 \lightrule
\textbf{ASQP Rest16} & 1264 & 316 & 544 & 2853 & 3040 & 2754 & 2279 \\
 & (1369, 62, 558) & (341, 23, 143) & (584, 40, 177) &  &  &  &  \\
 \lightrule
\textbf{ACOS Laptop} & 2934 & 326 & 816 & 4958 & 5378 & 4992 & 4958 \\
 & (2583, 227, 1364) & (279, 24, 137) & (716, 65, 380) &  &  &  &  \\
 \lightrule
\textbf{ACOS Res} & 1530 & 171 & 583 & 3110 & 3335 & 2967 & 3110 \\
 & (1656, 95, 733) & (180, 12, 69) & (668, 44, 205) &  &  &  &  \\  
 \lightrule
\textbf{EduRABSA} & 4000 & / & 2500 & 16884 \textsuperscript{\textcolor{myblue}{*}} & 26533 & 18148 & 10510 \\
 & (9581, 1713, 5206) & / & (5994, 1049, 3494) &  &  &  &  \\ 

\bottomrule

\end{tabular}
}

\caption*{
\begin{minipage}{\textwidth}
\footnotesize
\setlength{\baselineskip}{1.2\baselineskip}
    \hangindent=1em \hangafter=1 \hspace{0.05\textwidth} \textsuperscript{1} ASQP, ACOS, and ASQE all extract quadruplets, ACOS and ASQE (ours) also include implicit aspects and opinions \cite{compound_task_survey_2025}. \par
    
    \hangindent=1em \hangafter=1 \hspace{0.05\textwidth} \textsuperscript{2} EduRABSA: Quadruplets with implicit aspect = 2,456. \par

\end{minipage}
}

\end{table}


\renewcommand{\arraystretch}{1.25}

\begin{table}[!htbp]
\centering
\footnotesize

\caption{Aspect, Opinion Term Length Comparison: EduRABSA vs. Benchmark ASQE Datasets from \cite{compound_task_survey_2025}. \\``R'' and ``L'' in dataset names refer to ``Res'' and ``Lap'', respectively.} \vspace{3pt}
\label{table_dataset_text_length}

\resizebox{1\linewidth}{!}{%
\begin{tabular}{l|rrrrr|rrrrr}

\hline

\textbf{Length} & \multicolumn{5}{c|}{\textbf{Aspect}} & \multicolumn{5}{c}{\textbf{Opinion}} \\
(\#Tokens) & EduRABSA & ASQP 15R & ASQP 16R & ACOS L & ACOS R & EduRABSA & ASQP 15R & ASQP 16R & ACOS L & ACOS R \\

\hline
1 & 17,190 & 1,977 & 2,609 & 4,395 & 2,926 & 5,734 & 1,933 & 2,550 & 5,539 & 3,320 \\
2 & 6,880 & 322 & 423 & 1,111 & 426 & 6,083 & 277 & 360 & 125 & 208 \\
3 & 1,461 & 115 & 152 & 144 & 175 & 3,816 & 114 & 151 & 76 & 74 \\
4 & 705 & 41 & 59 & 94 & 78 & 2,998 & 87 & 120 & 24 & 24 \\
5 & 387 & 28 & 34 & 21 & 34 & 2,315 & 45 & 64 & 4 & 22 \\
6 & 200 & 6 & 7 & 6 & 8 & 1,802 & 22 & 27 & 4 & 8 \\
7 & 108 & 3 & 5 & 0 & 6 & 1,270 & 6 & 8 & 0 & 4 \\
8 & 83 & 1 & 1 & 1 & 4 & 955 & 4 & 4 & 0 & 0 \\
9 & 46 & 1 & 2 & 1 & 2 & 640 & 2 & 3 & 1 & 0 \\
10 & 37 & 0 & 0 & 0 & 0 & 450 & 3 & 5 & 0 & 0 \\
11-20 & 40 & 2 & 3 & 0 & 2 & 1,049 & 3 & 3 & 0 & 1 \\
21-30 & 2 & 0 & 0 & 0 & 0 & 26 & 0 & 0 & 0 & 0 \\
31-40 & 0 & 0 & 0 & 0 & 0 & 1 & 0 & 0 & 0 & 0 \\
\lightrule
\textbf{TOTAL} & \textbf{27,139} & \textbf{2,496} & \textbf{3,295} & \textbf{5,773} & \textbf{3,661} & \textbf{27,139} & \textbf{2,496} & \textbf{3,295} & \textbf{5,773} & \textbf{3,661} \\
\hline
\end{tabular}%
}

\end{table}


\newpage
\section{Related Work}

Table \ref{table_eduabsa_studies} summarises the ABSA tasks and dataset details of the 15 education review ABSA studies published since 2008 from the two systematic searches mentioned in the previous section. In terms of ABSA task coverage, over 92.3\% (N=12) formulated the problem as chaining multiple tasks (i.e. AE and/or ACD, ASC) sequentially, either through a pipeline of task-specific modules (N=11) or in an end-to-end (E2E) unified system (N=1).  However, as the ABSA components are tightly intertwined and form context for each other, this multi-task approach is prone to error propagation, context isolation, and representational bottlenecks between interrelated components \cite{sk2_jointabsa, jointabsa, batch2_survey_absa, implicitOE_7}. To overcome these issues, an increasing number of recent E2E ABSA studies have adopted composite tasks such as AOPE, ASTE, and ASQE that can better capture the inter-task relations via shared context and
learning \cite{hua2024absa, batch2_survey_absa, jointabsa}. It is therefore important to include composite tasks in the education review ABSA datasets to enable this unified approach. 

With respect to dataset sources, many of these studies used Twitter posts and Coursera course reviews for their public accessibility. However, these reviews are often much simpler and shorter than the average course evaluations and general student survey open-ended comments from our experience. This could lead to downstream model robustness and evaluation validity issues that were raised on the benchmark SemEval datasets \cite{fei2026robustness, mams2019}. For public datasets, it is thus crucial to capture sufficient complexity and diversity to reflect real-life use cases. 

In addition, from the perspective of creating useful datasets for the entire education review ABSA domain, we identified two gaps.  First, none of these studies covered the OE task to extract the opinion expressions, while education review reporting commonly needs to examine certain raw text to understand the details. Second, a robust and realistic dataset should cover two less-studied ABSA subtasks that are prevalent in opinionated text, especially education reviews: Implicit Aspect Extraction (IAE) and Implicit Opinion Extraction (IOE) \cite{ref_2025_01, IAEsurvey2022, implicitOE_2, implicitOE_4, implicitOE_5}. 

An \textbf{implicit aspect} is absent from the text but can be inferred from the context \cite{IAEsurvey2022}, such as the inferred aspect \textit{``price''} in \textit{``The restaurant was expensive''}.   An \textbf{implicit opinion}\footnote{Some studies use the term ``implicit sentiment'', where we choose ``implicit opinion'' as it shows the direct relationship with explicit opinion and the (explicit) OE task.} is an expression that ``does not carry words with obvious sentiment tendencies, but rather expresses the sentiment implicitly with objective statements, and expresses itself through factual, metaphorical, or ironic expressions.'' \citep[p.~1]{implicitOE_1}. For example:

\begin{itemize}
    \item[] Explicit opinion: \textit{``The waiter was rude.''}  (Explicit negative opinion word ``rude'')
    
    \item[] Implicit opinion: \textit{``The waiter poured water on my hand and walked away.''} (No opinion word, but has clear negative sentiment through implicit opinion expression ``poured water on my hand and walked away'') \cite{implicitOE_2} 

\end{itemize}

Li et al. (2021) \cite{implicitOE_2} reported that around 27\% and 30\% of review entries in the most widely used ABSA benchmark datasets \cite{hua2024absa} -- SemEval 2014 Restaurant and Laptop \cite{semeval2014} contain implicit opinions. In our observation, implicit opinions are very common in student reviews, such as \textit{``Didn't learn a single thing.''}, \textit{``Anything I learned was self-taught.''}, \textit{``Would take again.''} (from \cite{waterloo_coursereviews, ratemyprofessor_teacherreviews}). In addition, teacher reviews often have sentences with only descriptions of staff behaviour that have clear contextualised sentiment orientation (e.g. \textit {``Never answers e-mails and cannot answer questions.''} (from \cite{ratemyprofessor_teacherreviews})). However, implicit opinion extraction has not been widely covered by existing ABSA studies and datasets \cite{implicitOE_2, implicitOE_5, hua2024absa}. 

We designed our dataset aiming to address the gaps identified above, while capturing the common domain categories and tasks suggested by the previous studies. As detailed in the next section, our dataset covers almost all single and composite ABSA tasks, and includes the extraction of both implicit aspects and opinions to facilitate the development of future solutions and tools with this capability. 

\renewcommand{\arraystretch}{1.5}

\begin{table*}[htbp]
\centering
\caption{Task and Dataset Summary of Education Review Domain ABSA Studies from 2008-2025}  

\caption*{\small{\textit{* Note 1: } 
These studies were from two systematic literature searches. Items \#1-12 were all the in-domain studies from a systematic review of 519 ABSA studies from 2008-2023 \cite{hua2024absa}. Items \#13-15 were from the 502 results of Search 2 (details below) for 2024-2025 ABSA publications from three main databases (*), among which three 2024 in-domain ABSA studies were excluded due to no information on ABSA method (N=2) and experiment (N=1).}}  

\caption*{\small{\textit{* Note 2: } For Search 2, we searched ACM Digital Library, IEEE Xplore, and Springer Link with a 2024-2025 publication year filter and the following keywords: \textit{(``aspect-based sentiment analysis'' OR ``aspect based sentiment analysis'') AND (``student'' OR ``course'' OR ``teaching'' OR ``teacher'' OR ``education'' OR ``MOOC'')}.}}  

\label{table_eduabsa_studies}

\scriptsize 
\setlength{\tabcolsep}{3pt} 
\begin{tabularx}{\linewidth}{@{}
>{\raggedright\arraybackslash}p{0.01\linewidth} 
>{\raggedright\arraybackslash}p{0.09\linewidth} 
>{\raggedright\arraybackslash}p{0.07\linewidth} 
>{\raggedright\arraybackslash}p{0.1\linewidth} 
>{\raggedright\arraybackslash}p{0.06\linewidth} 
>{\raggedright\arraybackslash}p{0.19\linewidth} 
>{\raggedright\arraybackslash}p{0.07\linewidth} 
>{\raggedright\arraybackslash}p{0.08\linewidth} 
>{\raggedright\arraybackslash}p{0.25\linewidth}@{} 
}
\toprule

\textbf{\#} & \textbf{Title} & \textbf{Domain} & \textbf{ABSA tasks} & \textbf{Dataset Public} & \textbf{Datasets Used} & \textbf{Dataset Language} & \textbf{Sentiment Labels}\textsuperscript{\textcolor{blue}{1}} & \textbf{Aspect Categories} \\ 

\midrule

1 & Sivakumar \& Reddy (2017) \cite{edu_11_tweet} & Student Tweets \& posts & AE, ASC & No & Twitter tweets and online posts (Size unclear, original, unlabelled) & English & P, Nu, Ng & (N = 7) Teaching, Placement, Facilities, Sports, Organizing events, Fees, Transport \\
2 & Sindhu et al. (2019) \cite{edu_1_hmm} & Teacher review & ACD, ASC & No & Students feedback of Sukkur IBA University (N = 5000+, original, manually labelled) & English & P, Nu, Ng & (N = 6) Teaching pedagogy, Behavior, Knowledge, Assessment, Experience, General \\
3 & Chauhan et al. (2019) \cite{Edu_4_174} & Student Tweets & AE, ACD, ASC & No & Student tweets (N = 1000, original, unlabelled) & Thai & P, Nu, Ng, C & N/A, Cluster-based \\
4 & Soe \& Soe (2019) \cite{edu_10_ontology} & Course, Teacher, Facility & AE, ASC & No & UCST student feedback (size unclear, original); Manually created domain aspect ontology & English & P, Ng & N/A \\
5 & Kastrati et al. (2020) \cite{edu_3_wealysupervised} & MOOC review & ACD, ASC & No & Coursera course review (N = 104999, original, partially manually labelled) & English & P, Nu, Ng & (N = 9) Course (Content, Structure, General), Instructor (Knowledge, Skill, Experience, Interaction), Assessment, 
Technology \\
6 & Kastrati et al. (2020) \cite{edu_6_idcnn} & MOOC review & ACD, ASC & No & Coursera course review (N = 21940, original, manually labelled) & English & P, Nu, Ng & (N = 5) Instructor, Content, Structure, Design, General \\
7 & Alassaf \& Qamar (2020) \cite{edu_12_hybrid} & Student Tweets & AE, ASC & No & Tweets related to the institute (N = 8023, original, manually labelled) & Arabic & P, Non-Ng & (N = 9) Teaching, Environment, Electronic services, Staff affairs, Academic affairs, Activities, Student affairs, Higher education, Miscellaneous \\
8 & Wehbe et al. (2021) \cite{edu_8_uae} & Student Tweets & ACD & Yes (unlabelled) & UAE Twitter Dataset (Task dataset, N = 171873, original, unlabelled); COVID-19 Tweets (manually labelled) \textsuperscript{\textcolor{blue}{2}} & English, Arabic & N/A & (N = 5) Educational rights, Financial security, Job security, Safety, Death \\
9 & Hussain et al. (2022) \cite{edu_2_aspect2labels} & Course, Teacher, University review & ACD, ASC & No & Students reviews from  NCBA\&E University (N = 5767, original, unlabelled) & English & P, Nu, Ng & (N = 11) Teacher (Subject matter knowledge, Experience, Behavior, General), Course (Objective \& goals, Course content, Assessment, General), University (Environment, Policy, General) \\
10 & Ren et al. (2022) \cite{edu_5_autoscoring} & Teacher review & ASC & On request & Chinese Education Bureau teaching evaluation (N = 4483, original, manually labelled) & Chinese & P, Ng & (N = 9) Teacher quality, Teacher image, Teaching method, Teaching content, Teaching ability, Teaching attitude, Teaching effect, Teacher-student relationship, Classroom atmosphere \\
11 & Almatrafi \& Johri (2022) \cite{edu_7_suggestion_mining} & MOOC review & AE; ASC & N/A & The Stanford MOOCPosts dataset (used N = 906, manually labelled for ABSA) & English & P, Nu, Ng & (N = 5) Assessments, General (Course), Instructional staff, Material, Others \\
12 & Edalati et al. (2022) \cite{edu_9_mooc} & MOOC review & ACD; ASC & No & Coursera Course Review (N = 21940, from \cite{edu_6_idcnn}) & English & P, Nu, Ng & (N = 5) Instructor, Content, Structure, Design, General \\
13 & Dissanayake \& Fernando (2025) \cite{edu_2025_02} & course review & Review level categorisation & No & Udemy course review (N = 10K, original, manually labelled); 100K Coursera Course Review (N = 100K, unlabelled) \textsuperscript{\textcolor{blue}{3}} & English & P, Ng & (N = 7) Instructor, Content, Assignment, Interactions, Tech issues, Timeliness, General \\
14 & Gul et al. (2025) \cite{edu_2025_03} & Course \& teacher review & ACD, ASC & On request & Student review data from National Textile University  (N = 11747, original, manually labelled) & English & P, Ng & (N = 10) Assessment, Behavior, Experience, Knowledge, Teaching skills, Teacher general, Course structure, Course material,  Lab/Practical, Course general \\
15 & Marir et al. (2025) \cite{edu_2025_06} & course review & ACD, ASC & Yes (unlabelled) & Udemy Courses dataset (used = 104712, unlabelled) \textsuperscript{\textcolor{blue}{4}} & English & P, Nu, Ng & (N = 4) Course, Instructor, Assessment, Technology \\ 

\bottomrule

\end{tabularx}

\caption*{
\begin{minipage}{\textwidth}
\scriptsize 
\setlength{\baselineskip}{1.2\baselineskip}
\hangindent=2.5em \hangafter=1
\textsuperscript{1} P: Positive; Nu: Neutral; Ng: Negative; C: Conflict.\par
\hangindent=0.75em \hangafter=1
\textsuperscript{2} For obtaining aspect categories. Unlabelled version from https://www.kaggle.com/datasets/gpreda/covid19-tweets. \par
\hangindent=0.75em \hangafter=1
\textsuperscript{3} Unlabelled original from https://www.kaggle.com/datasets/septa97/100k-courseras-course-reviews-dataset.  \par
\hangindent=0.75em \hangafter=1
\textsuperscript{4} Unlabelled original from https://www.kaggle.com/datasets/hossaingh/udemy-courses. \par
\end{minipage}
}

\end{table*}

\renewcommand{\arraystretch}{1}

\newpage

\section{Dataset Details}\label{sec3_dataset_details}

The \textbf{Education Review ABSA (EduRABSA) dataset} consists of 6,500 entries of public tertiary student review text in the English language released in 2020-2023 on courses (``course review'', N=3,000), teaching staff (``teacher review'', N=3,000), and university (``university review'', N=500). Each text entry was manually annotated to extract a list of all relevant ``aspect, opinion, category, sentiment'' quadruplets and assign an overall sentiment label for the entire entry. The aspect and opinion terms were extracted as verbatim consecutive words. The category labels follow the ``entity-attribute'' two-level structure from SemEval 2015-2016 \cite{semeval2015, semeval2016} and include ``Course'', ``Staff'', and ``University'' (the entities) and their main attributes. The quadruplet-level and review-level sentiment categories are ``Positive'', ``Neutral'', and ``Negative''. The quadruplet extraction captured both explicit and implicit aspect- and opinion-terms \cite{implicitOE_1, implicitOE_2}, as well as the multi-aspect multi-sentiment relations on the same opinion term \cite{mams2019}. 

\begin{figure}[!hbp]
  \centering
  \includegraphics[width=0.75\linewidth]{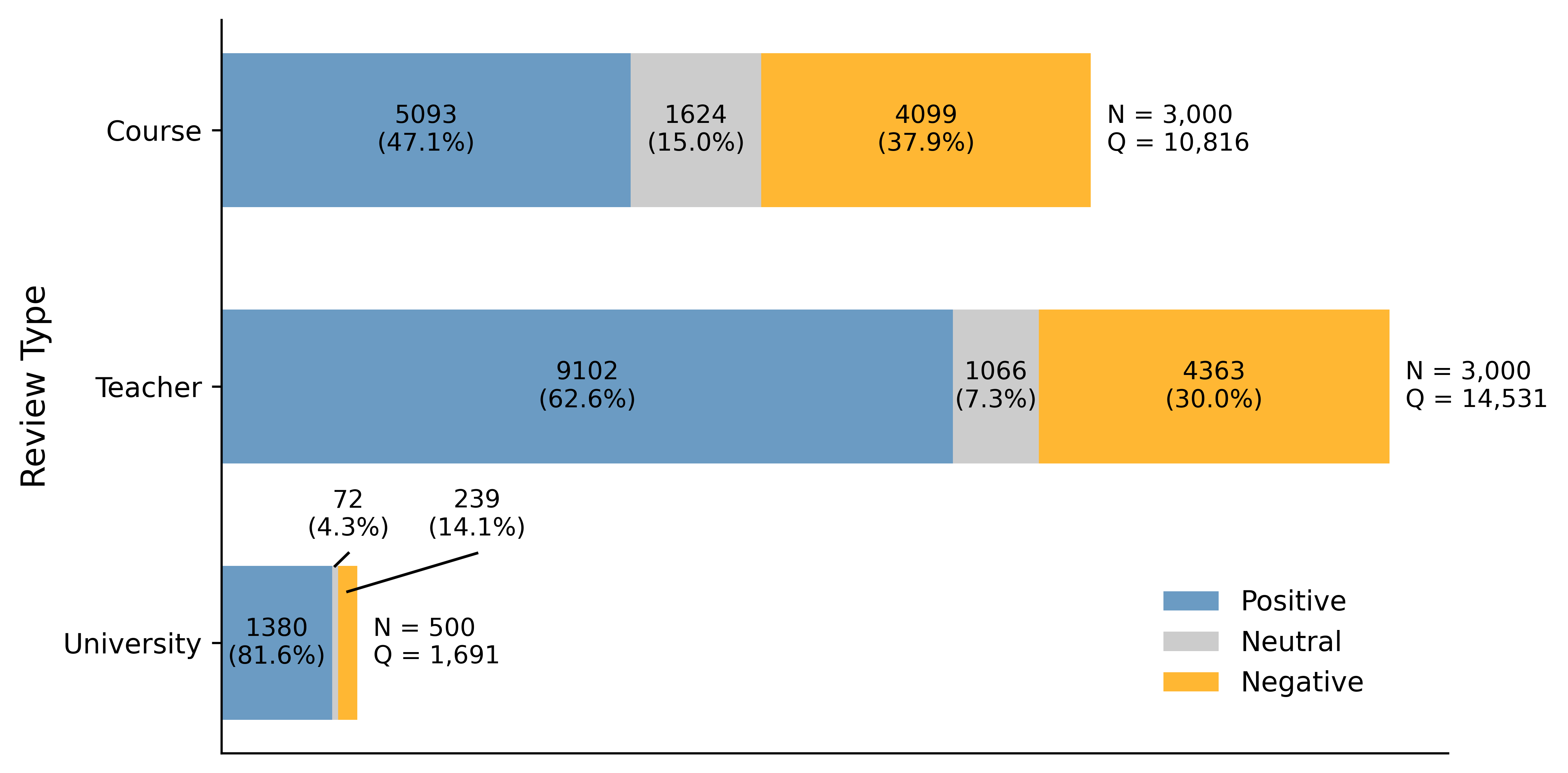}
  \caption{Counts and percentages of review entries (N) and quadruplets (Q) across review types and aspect sentiment categories in the EduRABSA dataset}
  \label{fig_dataset_info}
\end{figure}

Tables~\ref{table_dataset_stats}--\ref{table_dataset_text_length} detail the key dataset statistics compared with the commercial-domain benchmark counterparts. Figure \ref{fig_dataset_info} visualises the number of extracted quadruplets and aspect sentiment labels across the three review types. We also provide a few examples of review text and annotations in the EduRABSA dataset in Table \ref{table_dataset_example} in the Appendix. The mixture of mentions of course and staff attributes in either course or teacher review is fairly common in this domain. 

The sections below provide details of the data sources, sampling strategy, and annotation method. The overall process is illustrated in Figure \ref{fig_overall_image}.

\subsection{\textbf{Source Review Data}} \label{A1_source_data}

To create the EduRABSA dataset, we sampled 6,500 pieces of review text from three open-source, public-domain licensed tertiary student reviews datasets listed in
 Table \ref{table_dataset_1}. The course and teacher review datasets were released before the prevalence of generative LLMs (e.g. ChatGPT \cite{refChatGPT}, released in November 2022) and are therefore likely to be free from LLM-generated content. 

\renewcommand{\arraystretch}{1.5}

\begin{table*}[!htbp]
\footnotesize
\caption{Information of the unannotated source data of the Edu Reviews ABSDA dataset}
\label{table_dataset_1}

\begin{tabularx}{\textwidth}{
>{\raggedright\arraybackslash}p{0.13\linewidth} 
>{\raggedright\arraybackslash}p{0.35\linewidth} 
>{\raggedright\arraybackslash}p{0.1\linewidth} 
>{\raggedright\arraybackslash}p{0.1\linewidth} 
>{\raggedleft\arraybackslash}p{0.08\linewidth} 
>{\raggedleft\arraybackslash}p{0.08\linewidth}}
\toprule
\textbf{Review Type} & \textbf{Dataset Name} & \textbf{Publish Year} & \textbf{Licence} & \textbf{Total Entries} & \textbf{Sampled (N=6,500)} \\
\midrule
Course review & Course Reviews University of Waterloo  \cite{waterloo_coursereviews} & October 2022 & CC0: Public Domain & 14,810 & 3,000 \\
Teacher review & Big Data Set from RateMyProfessor.com for Professors' Teaching Evaluation  \cite{ratemyprofessor_teacherreviews} & March 2020 & CC BY 4.0 & 19,145 & 3,000 \\
University review & University of Exeter Reviews  \cite{exeter_universityreviews} & June 2023 & CC0: Public Domain & 557 & 500 \\
\bottomrule
\end{tabularx}
\end{table*}

\renewcommand{\arraystretch}{1}


\subsection{\textbf{Dataset Sampling}} \label{A2_source_sampling}

Our sampling strategy focused on realistic and representative sample distributions per review type, so that the output dataset can be used for multiple purposes, such as drawing a rebalanced subset for model training/testing, conducting fine-grained insight analysis, and providing distributional statistics for data augmentation and synthesis. We applied stratified sampling to each of the three data sources (i.e. review types) based on two criteria: 1) review text length, and 2) review rating score. 

We represented review text length using token count\footnote{We used the NLTK \cite{nltk} word\_tokenize library: \url{https://www.nltk.org/api/nltk.tokenize.word_tokenize.html}} binned in intervals of 20. The rating scores were grouped into ``low / medium / high'' categories as follows: For Course Reviews, we used the Boolean rating ``liked the course", with 0 mapped to ``low'' and 1 mapped to ``high''. The overall ratings for Teacher and University reviews both range from 1.0 to 5.0, with values in the range [1.0, 2.5] mapped to ``low'', (2.4, 4] to ``medium'', and (4.0, 5.0] to ``high''. 

Figures \ref{fig_dataset_1} through \ref{fig_dataset_4} in Appendix B show the source and sample dataset distribution comparison.


\subsection{\textbf{Sentiment and Aspect Categories}} \label{A3_categories}

We determined the sentiment polarity and aspect category labels before data annotation.  We chose the sentiment labels \textit{Positive}, \textit{Neutral}, and \textit{Negative} for consistency with most ABSA studies we reviewed. For the aspect category labels, we used each review type as the main category, and developed the sub-categories based on a pilot annotation of 200 review entries, as well as the common categories reported in the past education review ABSA studies detailed in Table \ref{table_eduabsa_studies}. The aspect categories are listed below, while their definitions with some examples are provided in Table \ref{table_category_mapping}.

\begin{itemize}
    \item Course (Content, Learning activity, Assessment, Workload, Difficulty, Course materials, Technology \& tools, Overall);
    \item Staff (Teaching, Knowledge \& skills, Helpfulness, Attitude, Personal traits, Overall);
    \item University (Cost, Opportunities, Programme, Campus \& facilities, Culture \& diversity, Information services, Social engagement \& activities, Overall)
\end{itemize}


\subsection{\textbf{Data Annotation}} \label{A4_annotation}

We manually annotated the sampled dataset using the ASQE-DPT tool (introduced in Section \ref{sec4_dpt}), using the following three-stage process (illustrated in Figure \ref{fig_overall_image}).

\textbf{1. Protocol development with pilot annotation}\quad We developed an initial protocol following that of the SemEval 2014-2016 ABSA datasets \cite{semeval2014, semeval2015, semeval2016} due to their wide usage and impact on the ABSA community \cite{hua2024absa}. We then carried out a pilot annotation on a random sample of 200 text entries to test the protocol and identify confusions. The pilot sample size was chosen for both feasibility and sufficient size to show diversity across review types. After the pilot, we discussed the identified ambiguities and refined the protocol accordingly. We also finalised the list of aspect categories to align with common tertiary student review use cases (e.g. \cite{edu_2025_03, edu_10_ontology, edu_11_tweet}) and the aspect categories used in prior education domain ABSA studies listed in Table \ref{table_eduabsa_studies}. This process resulted in the following six annotation rules:

    \begin{enumerate}
    
        \item[R1.] For each review text, first identify the opinion terms relevant to the main categories, then extract the target aspect terms of each opinion term. This is to avoid extracting aspect terms without opinion, such as a list of learning activities mentioned in a factual statement. 
        
        \item[R2.] Extract both explicit and implicit opinion terms. Implicit opinions are determined by whether, in the education context, an expression without opinion words conveys a clear sentiment relevant to the given categories. 
    
        \item[R3.] Extract both aspect terms and opinion terms as verbatim consecutive words. Aspect terms with conjunction or disjunction should be extracted as the maximum phrase following the SemEval approach to avoid ambiguity \cite{semeval2015}, e.g. ``this professor'' instead of ``professor'' as we found the review text often mentions multiple staff or courses. In addition, the extraction boundary of both aspect and opinion terms should cover enough content for readability and clarity. 
    
        \item[R4.] Each category label is a combination of one ``main category'' (i.e. entity) and one of its ``sub-category'' (i.e. entity attribute) from Table \ref{table_category_mapping}. E.g. ``Course - Assessment'' or ``Staff - Attitude''. Only assign one category label to each aspect-opinion pair. The categorisation should consider both aspect and opinion terms as context to avoid ambiguity between main categories, or from implicit aspect terms. 
    
        \item[R5.] If within the sentence the direct opinion target is a pronoun (e.g. ``this'', ``he/she''), refer to the adjacent sentence for the direct aspect term such as course name, staff name, or staff reference (e.g. ``this professor''). If none is available, use a pronoun instead. In the case of implicit aspects (e.g. if the entire course review text is ``It's a waste of time!''), use ``null'' to represent the implicit aspect term (course). 
    
        \item[R6.] Use ``neutral`` for very weak or mildly positive / negative sentiment \cite{semeval2015}. For review-level overall sentiment, ``neutral`` also represents mixed sentiment at the quadruplet level, e.g. ``A fairly useless and hard course. The concepts and labs are pretty cool though, which makes this course bearable''. 
    
    \end{enumerate}

\textbf{2. Manual annotation}\quad The annotation was implemented by the first author, who is an experienced domain expert in tertiary student evaluations and surveys. During the annotation process, we encountered a number of additional ``edge case'' aspects that were challenging to categorise largely due to the highly contextualised nature of teacher behaviour and course aspects. These cases were discussed and resolved with the other authors, one of whom is an education subject expert. The annotator then updated the annotation rules and previous relevant annotations accordingly. We share some of these ``edge cases'' and our decisions in Appendix \ref{appendix_B_annotation_notes}. 

\textbf{3. Final review}\quad After annotating all 6,500 entries, the annotator manually reviewed all annotations against the protocol and notes and made necessary revisions to ensure consistency.

\section{ASQE-DPT: A slim offline annotation tool}\label{sec4_dpt}

\begin{figure*}[!htbp]
  \centering
  \includegraphics[width=0.85\linewidth]{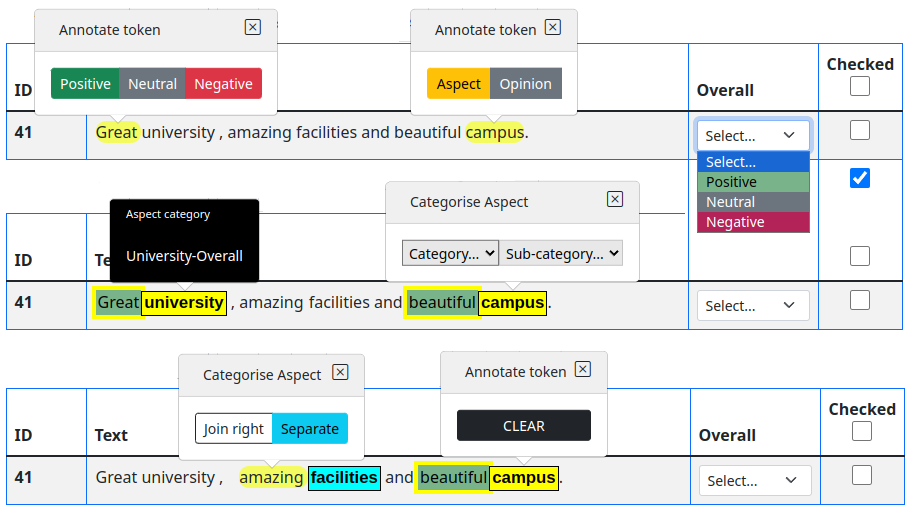}
  \caption{Screenshot of the ASQE-DPT Annotation Tool}
  \label{fig_dpt_screenshot}
\end{figure*}

ASQE-DPT\footnote{Available at \url{https://github.com/yhua219/edurabsa_dataset_and_annotation_tool}} is a manual text review and ABSA annotation tool we adapted from the PyABSA Data Preparation Tool (DPT)\footnote{\url{https://github.com/yangheng95/ABSADatasets/tree/v2.0/DPT}}. It inherited the original DPT's architecture, which is a self-contained HTML file with JavaScript and Vue components that can be used offline and directly on a browser without installation. It follows the original DPT's simple interface that displays review text in an interactive table form and only relies on mouse clicks for data annotation. However, it significantly differs from the original DPT in terms of functionality and the types of ABSA subtasks covered. 

The original DPT only supports three tasks: AE and ASC for ABSA and review-level sentiment analysis (SA). It takes a user-uploaded CSV file of review text, displays it as a table, and allows users to assign sentiment labels to either the selected consecutive words or an entire review entry. The sentiment labels on words are displayed as highlight colours. It has a Save button that generates three consecutive output files, two for the annotation tasks and one for the DPT process file to record progress. 

ASQE-DPT extends the original tool by adding new features:

\textbf{1. Mouse-clicking annotation functions for comprehensive and more challenging ABSA tasks:} 
    
    \begin{itemize}
        \item OE and AOPE (Allows multi-aspect and multi-opinion pairing, and visualises linked pairs upon mouse hover over either of the annotated member terms),

        \item Implicit AE (Automatically marks an unpaired aspect (and unpaired opinion) as ``null''),
        
        \item ACD (Supports single and two-level user-defined aspect categories. The pre-defined categories can be updated in the HTML source file, and are displayed in a pop-over drop-down upon clicking on a labelled opinion/aspect term),

        \item ASTE and ASQE (The AOPE, ACD, and ASC annotations on the same aspect-opinion pair are automatically combined into ASTE and ASQE annotations).

        \item A 'Clear' option to undo annotation.

    \end{itemize} 

\textbf{2. Auto-conversion of ASQE annotation into different task files.} The tool outputs a zip file containing five annotation files and one progress file that can be re-uploaded to review or continue annotation. The annotation files for all the tasks follow the corresponding format of the PyABSA platform  \cite{pyabsa} for easy downstream model training.

\textbf{3. Checking record and improved UI.} To support annotation review and record keeping, the new interface includes a checkbox column and a dynamic filename feature that records the number of rows marked as ``checked''.

\section{\textbf{Example Application}} \label{sec5_experiment}

We use a subsample of the new dataset to train models for two tasks. For ASC, we trained \textsc{FAST\_LSA\_T\_V2} from PyABSA \cite{pyabsa} using the example settings provided by the platform\footnote{Available at \href{https://github.com/yangheng95/PyABSA/blob/v2/examples-v2/aspect_polarity_classification/Aspect_Sentiment_Classification.ipynb}{https://github.com/yangheng95/PyABSA/.../aspect\_polarity\_classification/Aspect\_Sentiment\_Classification.ipynb}}. For ASQE, we trained \textsc{Tk-Instruct}\cite{tkinstruct} using the example script from PyABSA\footnote{Available at \href{https://github.com/yangheng95/PyABSA/blob/v2/examples-v2/aspect_opinion_sentiment_category_extraction/multitask_train.py}{github.com/yangheng95/PyABSA/.../aspect\_opinion\_sentiment\_category\_extraction/multitask\_train.py}}, and a \textsc{T5\_base} using the code and parameters from Li et al. (2023) \cite{two_stage_asqe_2023}. The two ASQE models were trained for 5 epochs. The training, validation, and test dataset sizes between EduRABSA and SemEval Restaurant 15/16 were all matched to 834 instances for training, 200 for validation, and 300 for testing. In addition, we also applied the same ASQE test set examples with pre-trained \textsc{GPT-4.1}, \textsc{Llama3.3-70B}, and \textsc{Llama3-8B} with zero-shot and 4-shot prompts provided in our code \footnote{\url{https://github.com/yhua219/edurabsa_dataset_and_annotation_tool}}.

\renewcommand{\arraystretch}{1.2}

\begin{table*}[!hbp]
\centering
\footnotesize

\caption{Test performance of models trained on EduRABASA subset for ASC/ASQE task}

\begin{tabularx}{0.9\linewidth}{
>{\raggedright\arraybackslash}p{0.15\linewidth}
>{\raggedright\arraybackslash}p{0.1\linewidth} 
>{\raggedright\arraybackslash}p{0.2\linewidth} 
>{\raggedleft\arraybackslash}p{0.1\linewidth} 
>{\raggedleft\arraybackslash}p{0.1\linewidth} 
>{\raggedleft\arraybackslash}p{0.08\linewidth} 
}

\toprule
\textbf{Dataset} & \textbf{Task} & \textbf{Model} & \textbf{Precision} & \textbf{Recall} & \textbf{F1-score} \\ \midrule
EduRABSA & ASC & FAST\_LSA\_T\_V2 & \multicolumn{2}{c}{(Accuracy) 0.79} & 0.58 \\

\lightrule
EduRABSA & ASQE & GPT-4.1\_fewshot & 0.25 & 0.28 & 0.26 \\
EduRABSA & ASQE & GPT-4.1\_zeroshot & 0.17 & 0.20 & 0.18 \\
EduRABSA & ASQE & Llama3.3-70B\_fewshot & 0.19 & 0.21 & 0.20 \\
EduRABSA & ASQE & Llama3.3-70B\_zeroshot & 0.13 & 0.15 & 0.14 \\
EduRABSA & ASQE & Llama3-8B\_fewshot & 0.14 & 0.15 & 0.14 \\
EduRABSA & ASQE & Llama3-8B\_zeroshot & 0.03 & 0.02 & 0.03 \\
\lightrule
EduRABSA & ASQE & T5\_base & 0.32 & 0.32 & 0.32 \\
EduRABSA & ASQE & Tk-Instruct\ & 0.26 & 0.26 & 0.26 \\
SemEval Rest15 & ASQE & T5\_base & 0.43 & 0.42 & 0.42 \\
SemEval Rest16 & ASQE & T5\_base & 0.57 & 0.57 & 0.57 \\ 
\bottomrule
\end{tabularx}
\end{table*}

\renewcommand{\arraystretch}{1}

The results for all pre-trained LLMs and \textsc{T5\_base} were evaluated using the latter's evaluation method. The two PyABSA models were evaluated using methods provided as part of their scripts. Our focus is on dataset comparison instead of the best training parameters. Overall, model performance on ASQE is lower than ASC, as the latter is a classification task on a given aspect, and the exact-string-matching evaluation method can be too harsh on ASQE for long aspect/opinion terms. As we have expected, all models performed worse on the EduRABSA data than the SemEval Restaurant datasets, proving our dataset to be more challenging in terms of length and the multi-aspect, multi-sentiment nature. On the other hand, the SemEval datasets do not cover implicit opinions, and were criticised for being overly simplified for model robustness \cite{mams2019, fei2026robustness}.


\newpage
\section{Conclusion and Future Work}\label{sec6_discussion}

ABSA is a promising solution for automated text mining of rich, fine-grained insights. The current public dataset domain skewness hinders its advancement and application in non-commercial domains, and particularly in the education review domain, which faces data sensitivity restrictions that greatly limit solution options, and research transparency and reproducibility \cite{hua2024absa}. Our work fills the gap by developing and sharing a publicly sourced, comprehensively annotated, realistically challenging ABSA dataset for course, teacher, and university reviews. The dataset aspect categories cover common education reporting use cases reflected by past studies; and the comprehensive, challenging ABSA task types enable more robust model training and evaluation beyond this domain. The ASQE-DPT annotation tool can further enable easy dataset creation and adaptation, particularly for low-resource areas and sensitive domains. 

\textbf{Limitations:} Despite our effort and domain expertise, the dataset annotation was conducted by a single annotator and thus lacks reliability statistics. As with any manual annotation, it also reflects intrinsically imperfect decisions on aspect and opinion boundaries and categories, particularly with the highly contextualised education reviews. The source reviews are only in the English language, and may not reflect concepts and events in later times, such as the topic of ``generative AI'' in education settings. The ASQE-DPT tool is currently targeted at PC monitor views and only supports mouse-based operations. 

\textbf{Future work:} By making both the dataset and the annotation tool open-source, we welcome the ABSA community to further explore and improve these resources. Future work could consider the following: 1) Developing more rigorous annotation rules that can support automatic annotation and quality evaluation by rule-based systems or generative LLMs. 2) Exploring more diverse ABSA approaches in this domain that further leverage the dataset, such as semi-supervised or contrastive learning with in-training data augmentation. 3) Extending the dataset into more languages (e.g. \cite{mabsa-2025} and categories or ABSA tasks. 4) Explore evaluation metrics for ASTE and ASQE tasks, particularly on more complex and longer texts that contain more units to extract, where classic exact string-matching becomes uninformative. We also hope that more research effort and resources could go to the public sectors and low-resource domains, so that the advancement of machine learning and natural language processing can benefit more areas and populations.

\renewcommand{\arraystretch}{1.5}

\begin{table*}[!b]
\small
\caption{Main Categories (entities) and Sub-Categories (attributes) in EduRABSA Dataset. }
\caption*{\footnotesize{The category label consists of a main category and one of its sub-categories connected by a hyphen, e.g. ``Course - Assessment'' or ``Staff - Attitude'' }}
\label{table_category_mapping}

\begin{tabularx}{\linewidth}{
>{\raggedright\arraybackslash}p{0.07\linewidth}
>{\raggedright\arraybackslash}p{0.2\linewidth}
>{\raggedright\arraybackslash}p{0.68\linewidth}
} 

\toprule
\textbf{Main Category} & \textbf{Sub-category} & \textbf{Example} \\
\midrule
Course     & Content                  & Course content, subject, topics \\
           & Learning activity        & Lectures, tutorials, workshops, lab sessions, field trips, group work, in-class activities \\
           & Assessment               & Assignments, homework, tests, exams, lab reports, quizzes  \\
           & Workload                 & Time and effort taken to complete the course activities or assessments  \\
           & Difficulty               & Difficulty level of course content (Assessment difficulty falls under the ``Assessment'' sub-category) \\
           & Course materials         & Lecture slides, handouts, coursebooks, textbooks, reading materials, class notes, lecture recordings, other teaching materials  \\
           & Technology \& tools      & Any technology, software and device used in the course \\
           & Overall                  & General opinion about the overall course, e.g. ``great course'', ``learned so much'', ``do not recommend''  \\
\midrule
Staff      & Teaching                 & Teaching method, behaviour, effectiveness, including things like ``going over materials after class'', ``giving a lot of assignments'', ``marks harshly'', and ``passionate about what he/she teaches''    \\
           & Knowledge \& skills      & Staff knowledge and skills  \\
           & Helpfulness              & Staff behaviours or attitudes that are perceived as being helpful/unhelpful, e.g. ``answers emails/questions'', ``willing to help'', ``ready to help''  \\
           & Attitude                 & Staff attitude towards the course and/or students, e.g. ``wants everyone to do well'', ``fair'', ``approachable'' \\
           & Personal traits          & Staff individual attributes, including physical features, intelligence, personality, accents, and general demeanour such as ``humorous'', ``very kind/sweet'' without specific context  \\
           & Overall                  & General opinion about the staff, e.g. ``he's/she's the best'', ``do not take''  \\
\midrule
University & Cost                     & Tuition and living cost at the university  \\
           & Opportunities            & Direct mention of ``opportunities'' \\
           & Programme                & Programme aspects mentioned in the university review context, e.g. course options, programme quality   \\
           & Campus \& facilities     & The environment and hardware of the university, including the relevant but non-university facilities such as ease of commute  \\
           & Culture \& diversity     & The university's culture and diversity in student and staff communities  \\
           & Information \& Services  & The provision and quality of information and services, e.g. counselling, help desk, campus cafe, gym.  \\
           & Social engagement \& activities & Campus social atmosphere, clubs/societies and activities, perceived social engagement with staff and peers  \\
           & Overall                  & Overall opinion about the university, e.g. ``the best uni'', ``waste of time''  \\
\bottomrule
\end{tabularx}
\end{table*} 

\renewcommand{\arraystretch}{1}



\clearpage
\newpage
\onecolumn

\appendix

\setcounter{figure}{0}
\setcounter{table}{0}

\renewcommand\thefigure{\thesection.\arabic{figure}}
\renewcommand\thetable{\thesection.\arabic{table}}
\renewcommand\theHtable{\thesection.\thetable}
\renewcommand{\thefigure}{\thesection.\arabic{figure}} 
\renewcommand{\theHfigure}{\thesection.\arabic{figure}} 

\begin{center}
{\large\bfseries APPENDICES}
\end{center}


\section{Sample Entries of the EduRABSA Dataset}\label{appendix_A_past_studies}


\renewcommand{\arraystretch}{1.15}

\begin{table*}[!hbp]
\small
\caption{Examples of the EduRABSA dataset review text and annotation (ASQE quadruplets (``Quad.''), and review-level sentiment label). Each row across columns shows a piece of review text and its review-level sentiment label; and the rows underneath list each extracted quadruplet containing one aspect term, one opinion term, their category label, and quadruplet-level sentiment label.}  

\label{table_dataset_example}

\begin{tabularx}{\textwidth}{
>{\raggedright\arraybackslash}p{0.11\textwidth}     
>{\raggedright\arraybackslash}p{0.15\textwidth}   
>{\raggedright\arraybackslash}p{0.22\textwidth}   
>{\raggedright\arraybackslash}p{0.25\textwidth}   
>{\raggedright\arraybackslash}p{0.15\textwidth} 
} 
\toprule
\textbf{\#} & \textbf{Aspect Term} & \textbf{Opinion Term} & \textbf{Category Label} & \textbf{Sentiment Label} \\ 
\midrule

1. Course review & \multicolumn{3}{>{\raggedright\arraybackslash}p{0.7\textwidth}}{``\textit{A fairly useless and hard course. The concepts and labs are pretty cool though, which makes this course bearable.}''} & Neutral \\
\midrule
Quad.1-1 & course & fairly useless & Course - Overall & Negative  \\
Quad.1-2 & course & hard & Course - Difficulty & Negative  \\
Quad.1-3 & concepts & cool & Course - Content & Positive  \\
Quad.1-4 & labs & cool & Course - Learning activity & Positive  \\
Quad.1-5 & course & bearable & Course - Overall & Neutral  \\
\midrule

2. Course review & \multicolumn{3}{>{\raggedright\arraybackslash}p{0.7\textwidth}}{``\textit{This class was challenging because the instructor was not able to express the materials clearly. Not to mention, he does not speak fluent English. In addition, it is even harder to understand the course materials, due to lack of samples and again fluent expression of the materials. Moreover, this class will require much practice and studying.}''} & Negative \\
\midrule
Quad.2-1 & This class & challenging & Course - Difficulty & Negative  \\
Quad.2-2 & instructor & not able to express the materials clearly & Staff - Teaching & Negative  \\
Quad.2-3 & instructor & does not speak fluent English & Staff - Personal traits & Negative  \\
Quad.2-4 & course materials & even harder to understand & Course - Course materials & Negative  \\
Quad.2-5 & course materials & lack of samples & Course - Course materials & Negative  \\
Quad.2-6 & this class & require much practice and studying & Course - Workload & Negative  \\
\midrule

3. Teacher review & \multicolumn{3}{>{\raggedright\arraybackslash}p{0.7\textwidth}}{``\textit{Took her for math 85 and math 90. Awesome teacher. She makes class entertaining and will stay late to help you understand the material. Take her for math!}''} & Positive \\
\midrule
Quad.3-1 & teacher & Awesome & Staff - Overall & Positive  \\
Quad.3-2 & teacher & makes class entertaining & Staff - Teaching & Positive  \\
Quad.3-3 & teacher & will stay late to help you understand the material & Staff - Helpfulness & Positive  \\
Quad.3-4 & teacher & Take her for math & Staff - Overall & Positive  \\
\midrule

4. Teacher Review & \multicolumn{3}{>{\raggedright\arraybackslash}p{0.7\textwidth}}{``\textit{He gives unclear instructions and unclear due dates. If you didn't do a project exactly how he wants it, he'll chew you out and degrade your art for 10-15 minutes in front of the class (yes this happens on multiple occasions). He's late about 75\% of the time (once even forgetting he had a class go teach and another bc he was looking at art).}''} & Negative \\
\midrule
Quad.4-1 & He & gives unclear instructions and unclear due dates & Staff - Teaching & Negative  \\
Quad.4-2 & He & late about 75\% of the time & Staff - Teaching & Negative  \\
Quad.4-3 & He & chew you out and degrade your art for 10-15 minutes in front of the class & Staff - Attitude & Negative  \\
\midrule

5. University review & \multicolumn{3}{>{\raggedright\arraybackslash}p{0.7\textwidth}}{``\textit{The University is great at sorting out career opportunities and the campus is great with very good facilities}''} & Positive \\
\midrule
Quad.5-1 & sorting out career opportunities & great at & University - Opportunities & Positive  \\
Quad.5-2 & campus & great & University - Campus \& facilities & Positive  \\
Quad.5-3 & facilities & very good & University - Campus \& facilities & Positive  \\

\bottomrule
\end{tabularx}
\end{table*}


\clearpage \newpage
\section{Annotation Notes on Edge Cases}\label{appendix_B_annotation_notes}

\definecolor{lightgray}{RGB}{245,245,245}
\setlength{\FrameSep}{10pt}

\begin{framed}

\colorbox{lightgray}{\parbox{\dimexpr\textwidth-2\fboxsep}{\centering\textbf{Classification Decisions of Edge Cases}}}

\medskip
\setlength{\tabcolsep}{6pt}
\begin{tabular}{@{}p{0.02\textwidth}p{0.45\textwidth}cp{0.42\textwidth}@{}}
01 & ``This course is lecture-only'' & $\rightarrow$ & (statement, not opinion) \\[0.5em]
02 & Staff assessment design, marking criteria & $\rightarrow$ & Staff-Teaching \\[0.5em]
03 & Assessment workload & $\rightarrow$ & Course-Workload \\[0.5em]
04 & Assessment difficulty and others & $\rightarrow$ & Course-Assessment \\[0.5em]
05 & Staff attitude and interaction with students & $\rightarrow$ & Staff-Attitude \\[0.5em]
06 & Staff cool, passionate, friendly, approachable & $\rightarrow$ & Staff-Attitude \\[0.5em]
07 & Staff appearance, accent, fun, humor, smart & $\rightarrow$ & Staff-Personal traits \\[0.5em]
08 & Peer interaction, group work, lectures, labs & $\rightarrow$ & Course-learning activity \\[0.5em]
09 & University campus & $\rightarrow$ & University-Campus \& facilities \\[0.5em]
10 & Course being fun, engaging, dry, useful & $\rightarrow$ & Course-Overall \\[0.5em]
11 & (Course review) ``learned a lot'', ``had a good time'' & $\rightarrow$ & Course-Overall \\[0.5em]
12 & ``bird course'' = easy course & $\rightarrow$ & Course-Difficulty \\[0.5em]
13 & Textbook, handbook, slides etc. & $\rightarrow$ & Course-Course materials \\[0.5em]
14 & ``great teacher'', ``best professor'' (on role) & $\rightarrow$ & Staff-Overall \\[0.5em]
15 & ``she's the best'', ``love her'', ``Take him'' & $\rightarrow$ & Staff-Overall \\[0.5em]
16 & Staff answer emails/questions & $\rightarrow$ & Staff-Helpfulness \\[0.5em]
17 & Readings & $\rightarrow$ & Course-Course material \\[0.5em]
18 & University nice people & $\rightarrow$ & University-Social engagement \& activities \\[0.5em]
19 & (University review) nice/helpful staff & $\rightarrow$ & Staff-Attitude \\[0.5em]
20 & Her class/His lecture & $\rightarrow$ & (Depends on focus: Course-Learning activity or Staff-Teaching) \\[1.5em]
21 & Willing to help/ready to help & $\rightarrow$ & Staff-Helpfulness \\[0.5em]
22 & Passionate about what he/she teaches & $\rightarrow$ & Staff-Teaching \\[0.5em]
23 & Wants everyone to do well & $\rightarrow$ & Staff-Attitude \\[0.5em]
24 & Sweetheart, sweet (without context) & $\rightarrow$ & Staff-Personal Trait \\[0.5em]
25 & You have to read and study/do the work & $\rightarrow$ & Course-Workload \\[0.5em]
26 & If you \ldots you will get A/you will be fine & $\rightarrow$ & Course-Difficulty or Course-Overall (aspect: course) \\[1.5em]
27 & I like going to class as he made it enjoyable! & $\rightarrow$ & Course-Learning activity; Staff-teaching (aspect: course) \\[1.5em]
28 & First half is easy, then it gets difficult & $\rightarrow$ & Course-Content \\[0.5em]
29 & Staff being fair & $\rightarrow$ & Staff-Attitude \\[0.5em]
30 & Staff go over materials with you after class & $\rightarrow$ & Staff-Teaching \\[0.5em]
31 & Gives a lot of homework/gives extra credit & $\rightarrow$ & Staff-Teaching \\
\end{tabular}

\end{framed}


\textbf{Detailed Explanations}

\begin{framed}
\textbf{Case 06 -- Staff-Attitude vs. Staff-Personal Trait}

``Staff-attitude'' captures interpersonal qualities, social approach, and manner of engagement, which ``cool'' directly addresses. ``Staff-personal trait'' refers more to inherent characteristics like personality type, which is more stable and less situational than attitude.
\end{framed}

\begin{framed}
\textbf{Case 26 -- ``If you ... you will get A/you will be fine'' }

This type of sentence is about the course. More particularly, describing the aspect of a course, such as overall outcome or difficulty. The specific subcategory under Course depends on the particular context and focus. 
\end{framed}

\begin{framed}
\textbf{Case 28 -- Course-Content vs. Course-Difficulty}

Sentences similar to ``First half is easy, then it gets difficult'' is classified as \textbf{Course-Content} because:
\begin{enumerate}
\item The comment describes how the subject matter develops over the course's duration
\item Difficulty here refers to a property of the content
\end{enumerate}

The phrase reveals how the course material transforms, becomes more complex, or introduces more advanced concepts---which is essentially a description of course content, with difficulty being an inherent attribute of that content.
\end{framed}

\begin{framed}
\textbf{Case 30 -- Staff-Helpfulness vs. Staff-Teaching}

The sentence ``He will go over materials with you after class if you need it'' is classified as \textbf{Staff-Helpfulness} rather than \textbf{Staff-Teaching}. The focus is on the availability of additional support and the staff member's willingness to assist the student after class, which is a form of help. \textbf{Staff-Teaching} would typically involve direct actions or instructions related to delivering course content during the class.
\end{framed}

\clearpage \newpage
\section{The Education Review ABSA (EduRABSA) Dataset Statistics}\label{appendix_C_figures}

\begin{figure*}[!htbp]
  \centering
  \includegraphics[width=0.90\linewidth]{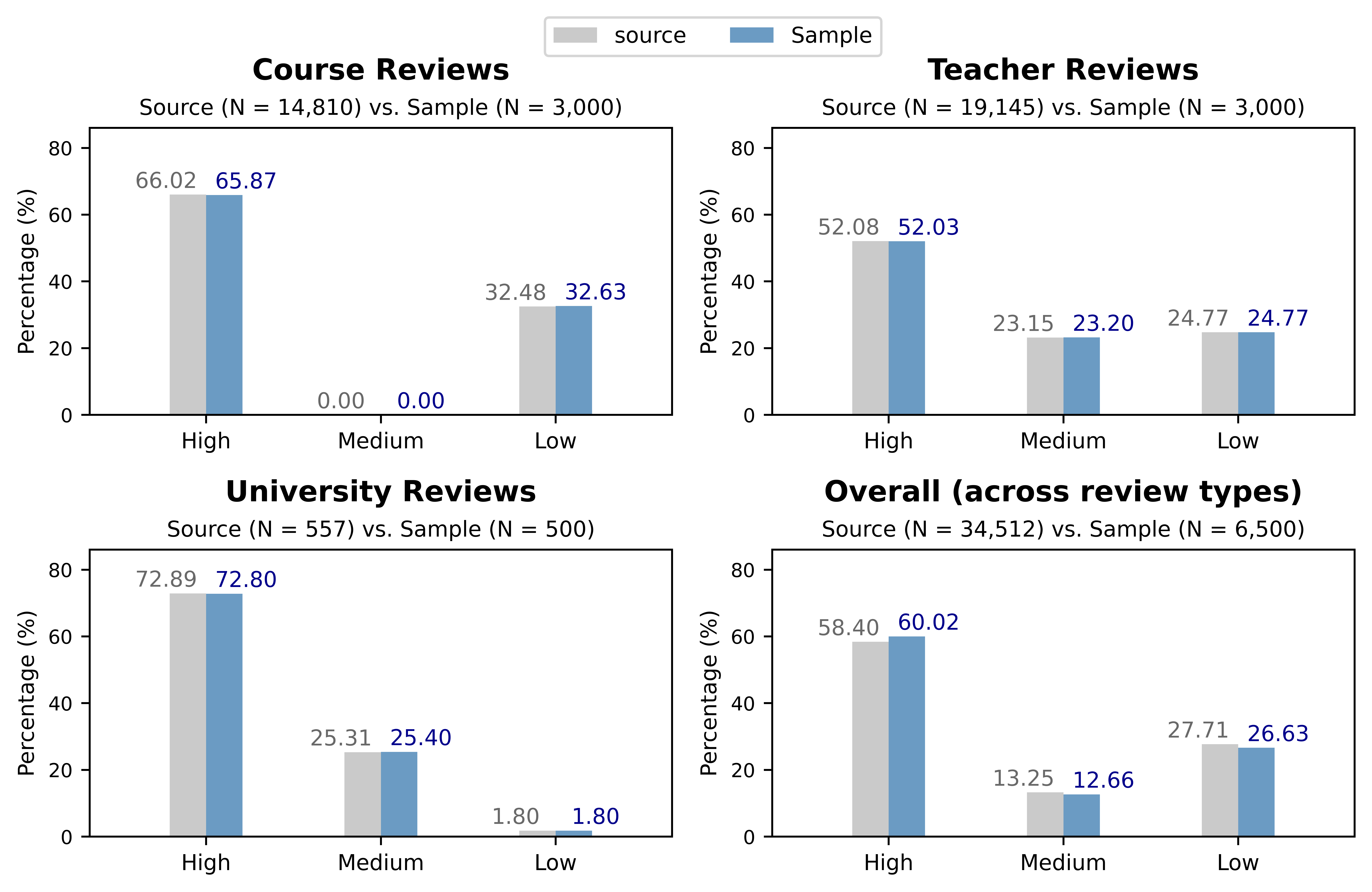}
  \caption{Percentage of review text entries (x-axis) by rating category (y-axis) for source and sampled datasets (left and right column in each pair) across review types. Note: For Course Reviews, we used the Boolean rating "liked the course", with 0 mapped to ``low'' and 1 mapped to ``high''. The overall ratings for Teacher and University reviews both range from 1.0 to 5.0, with values in the range [1.0, 2.5] mapped to ``low'', (2.4, 4] to ``medium'', and (4.0, 5.0] to ``high''. }
  \label{fig_dataset_2}
\end{figure*}

\begin{figure*}[!htb]
  \centering
  \includegraphics[width=0.85\linewidth]{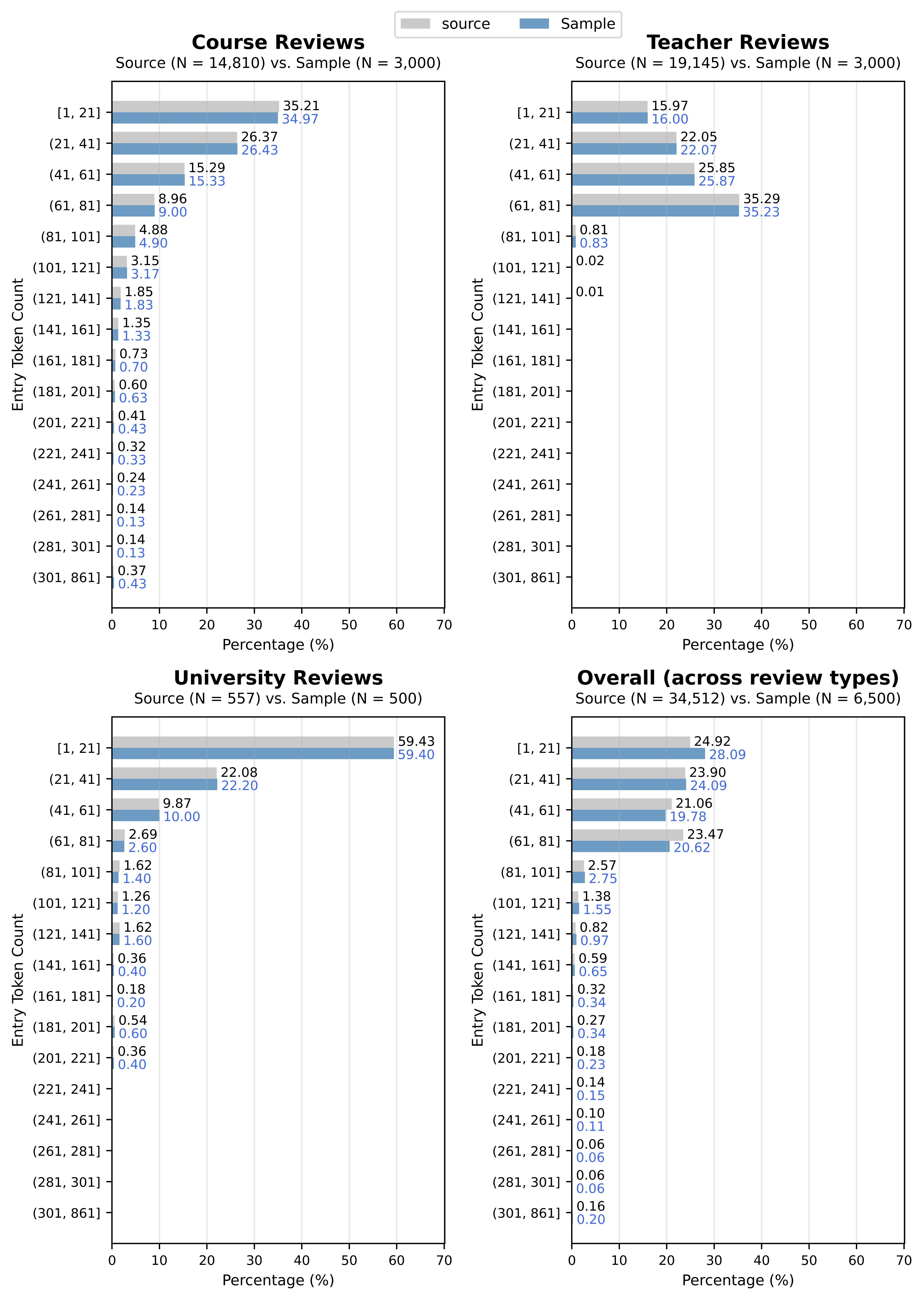}
  \caption{Percentage of review text entries (x-axis) by entry token count group (y-axis) for source and sampled datasets (top and bottom bars in each pair) across review types. For both source and sampled datasets, Course, Teacher, and University reviews have 86\%, 99\%, and 94\% entries within 80 tokens, and 91\%, 100\%, and 96\% entries within 100 tokens, respectively.}
  \label{fig_dataset_1}
\end{figure*}

\begin{figure*}[!htbp]
  \centering
  \includegraphics[width=\linewidth]{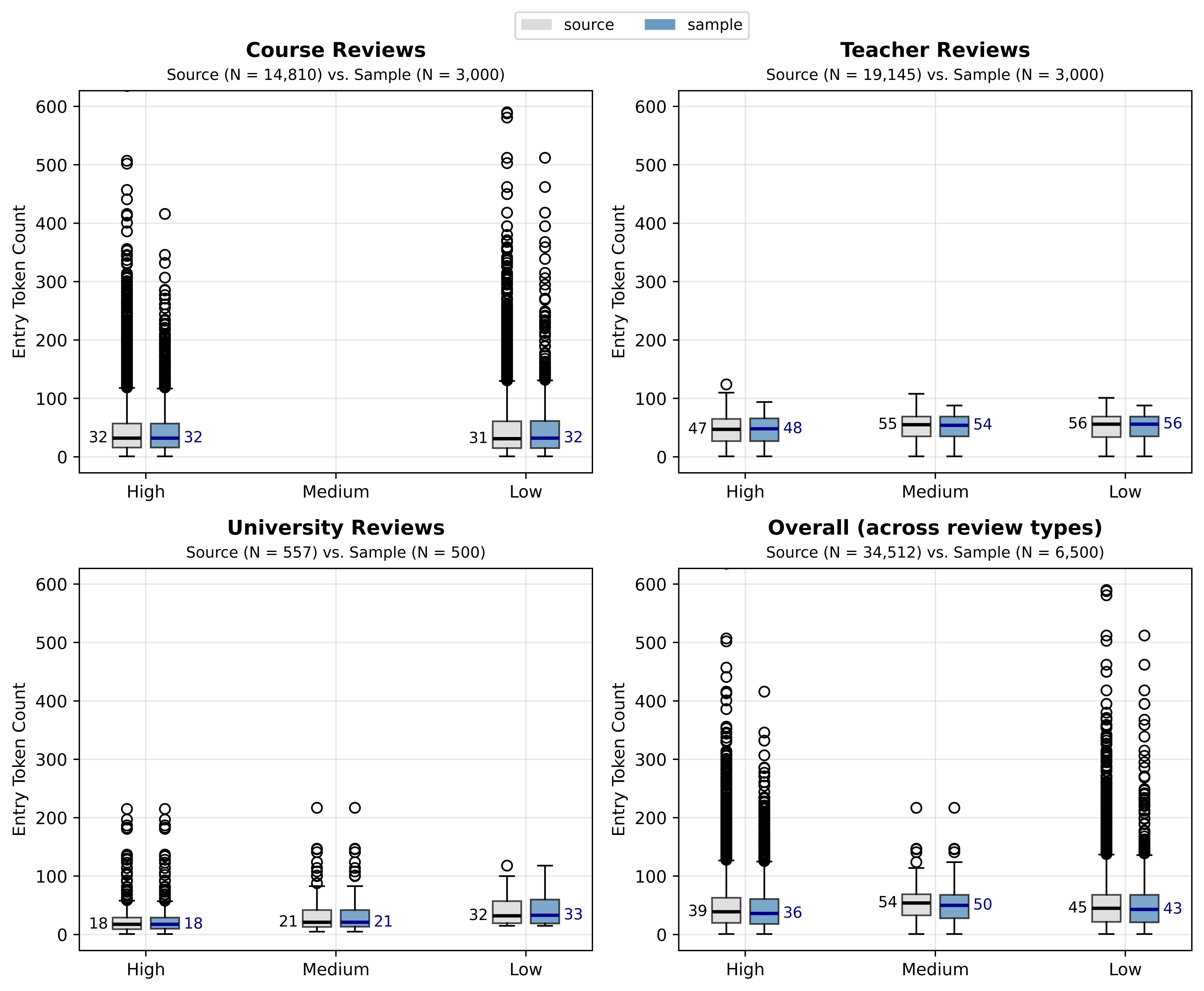}
  \caption{Summary of review text entry length (token count; y-axis) distribution by review rating category (x-axis) for source and sampled datasets (left and right components in each pair) across review types. Boxes show the middle 50\% of data (IQR; 25$^{th}$-75$^{th}$ percentiles) with labelled medians. Whiskers extend to data within 1.5×IQR; outliers appear as individual points. (Note: The Course Review source file has two entries of 635 and 856 tokens respectively and were excluded from sampling and this graph).}
  \label{fig_dataset_3}
\end{figure*}

\begin{figure*}[!htbp]
  \centering
  \includegraphics[width=0.78\linewidth]{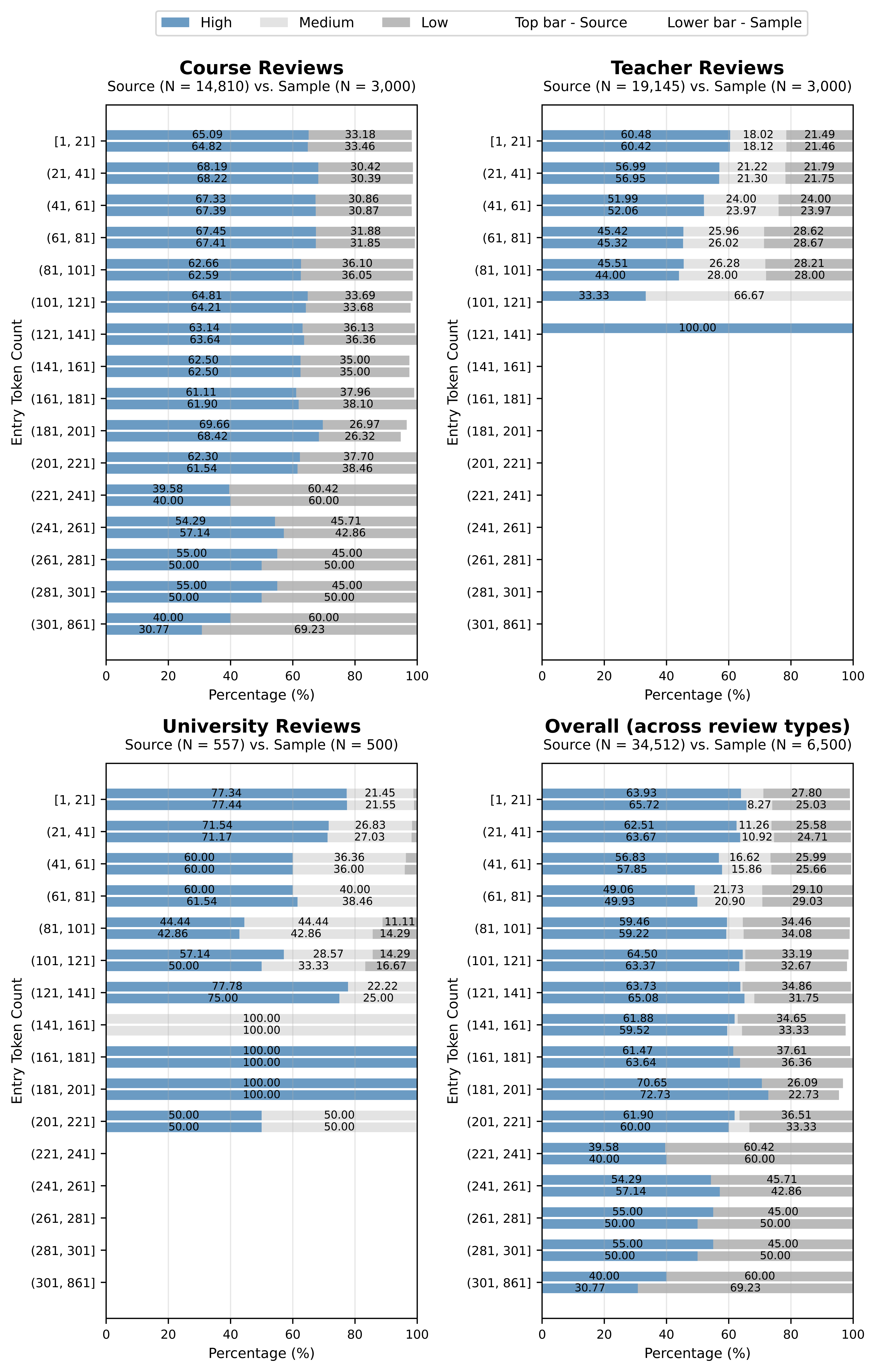}
  \caption{Percentage of review text entries (x-axis) by rating category (segments within bar) per entry token count group (y-axis) for source and sampled datasets (top and bottom bars in each pair) across review types. (Note: There were 221 original and 45 sampled Course Review entries without a rating score.)}
  \label{fig_dataset_4}
\end{figure*}


\clearpage
\bibliographystyle{unsrtnat}  

\bibliography{references}

\end{document}